\documentclass[10pt,twocolumn,letterpaper]{article}

\usepackage[pagenumbers]{iccv} % To force page numbers, e.g. for an arXiv version

\usepackage{pifont}% http://ctan.org/pkg/pifont

\usepackage{booktabs} % For formal tables
\usepackage{graphicx}
\usepackage{amsmath}
\usepackage{hyperref}
\usepackage{booktabs} % For formal tables
\usepackage{algpseudocode}
\usepackage{multirow}
\usepackage{subcaption} % For using subfigure environments
\usepackage{tcolorbox}
\definecolor{lightyellow}{RGB}{255,255,250}

\usepackage[ruled]{algorithm2e} % For algorithms

\SetAlFnt{\small}
\SetAlCapFnt{\small}
\SetAlCapNameFnt{\small}
\SetAlCapHSkip{0pt}

\title{Agentic 3D Scene Generation with Spatially Contextualized VLMs}

\author{
    Xinhang Liu\\
    HKUST
    \and
    Yu-Wing Tai\\
    Dartmouth College
    \and 
    Chi-Keung Tang\\
    HKUST 
}

\begin{document}

\twocolumn[
    {%
    \renewcommand\twocolumn[1][]{#1}%
    \maketitle
    \centering
    \vspace{-0.4cm}
    \includegraphics[width=\linewidth]{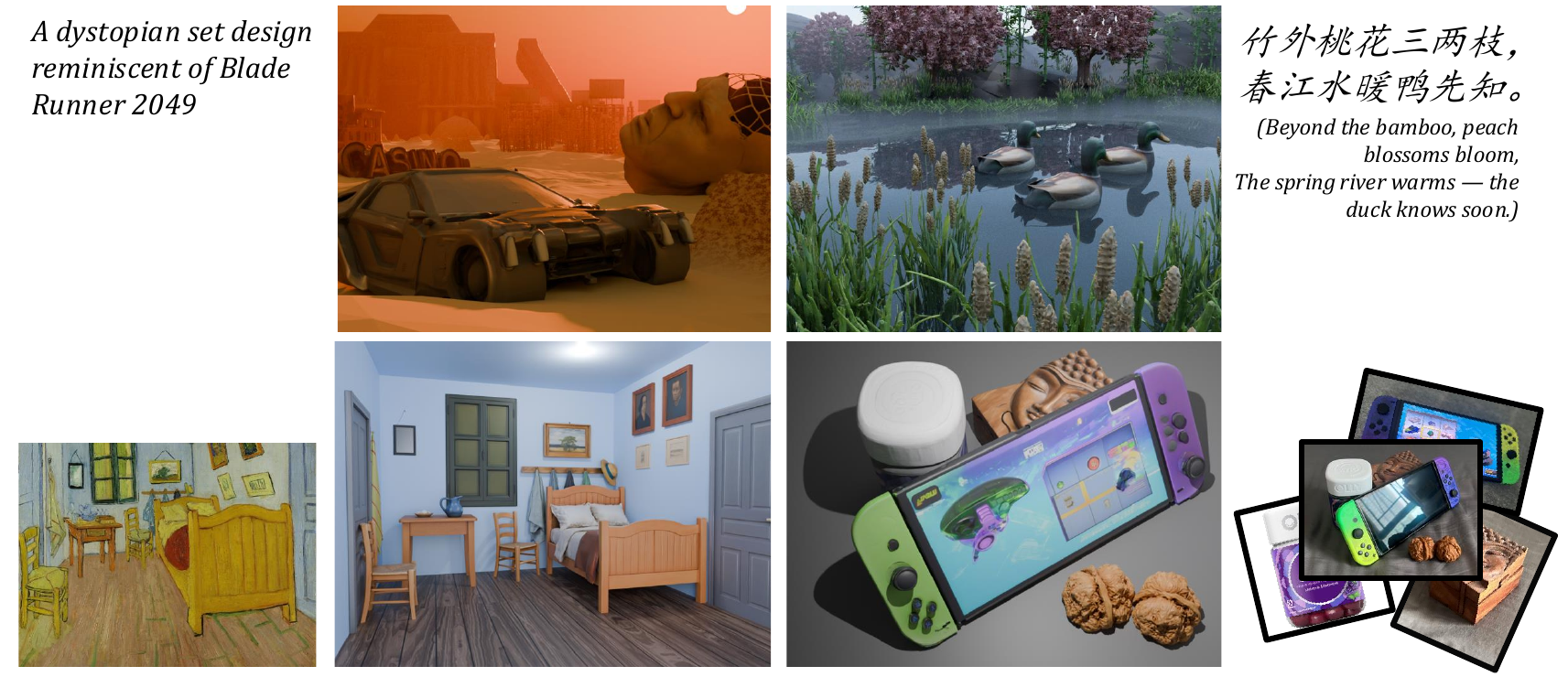}
    \captionof{figure}{\textbf{Spatially contextualized VLMs.} We propose a framework that equips VLMs with structured spatial context, enabling them to act as agents for 3D scene generation. Our approach supports diverse inputs—including text prompts, single images, and unstructured, unposed image collections—and produces coherent, semantically aligned 3D environments across a wide range of styles and settings. \label{fig:teaser}}
    \vspace{0.4cm}
}]
\begin{abstract}
Despite recent advances in multimodal content generation enabled by vision-language models (VLMs), their ability to reason about and generate structured 3D scenes remains largely underexplored. This limitation constrains their utility in spatially grounded tasks such as embodied AI, immersive simulations, and interactive 3D applications.
We introduce a new paradigm that enables VLMs to generate, understand, and edit complex 3D environments by injecting a continually evolving \emph{spatial context}. Constructed from multimodal input, this context consists of three components: \emph{a scene portrait} that provides a high-level semantic blueprint,  \emph{a semantically labeled point cloud} capturing object-level geometry, and \emph{a scene hypergraph} that encodes rich spatial relationships, including unary, binary, and higher-order constraints. Together, these components provide the VLM with a structured, geometry-aware working memory that integrates its inherent multimodal reasoning capabilities with structured 3D understanding for effective spatial reasoning.
Building on this foundation, we develop an agentic 3D scene generation pipeline in which the VLM iteratively reads from and updates the spatial context. The pipeline features high-quality asset generation with \emph{geometric restoration}, \emph{environment setup} with automatic verification, and \emph{ergonomic adjustment} guided by the scene hypergraph.
Experiments show that our framework can handle diverse and challenging inputs, achieving a level of generalization not observed in prior work. Further results demonstrate that injecting spatial context enables VLMs to perform downstream tasks such as interactive scene editing and path planning, suggesting strong potential for spatially intelligent systems in computer graphics, 3D vision, and embodied applications.
Project page: \url{https://spatctxvlm.github.io/project_page/}.

\end{abstract}

\section{Introduction}
Recent progress in multimodal content generation has demonstrated the impressive capabilities of large-scale vision-language models (VLMs) in interpreting and generating text, images, and even videos. Models such as GPT-4o have shown strong performance in tasks that require cross-modal reasoning, interactive grounding, and natural language understanding. Despite this progress, the ability of VLMs to reason about and generate structured 3D scenes remains largely underexplored. Unlike 2D content, structured 3D scenes (\Cref{fig:teaser}) impose additional demands such as maintaining spatial consistency, ensuring physical plausibility, and preserving semantic coherence.
This presents a fundamental limitation to the deployment of VLMs in spatially grounded applications such as embodied AI, robotics simulation, AR/VR content creation, and interactive environment design~\cite{kolve2017ai2, srivastava2022behavior, xiang2020sapien,szot2021habitat, yang2024holodeck}. Notably, these domains demand structured awareness of spatial geometry to support coherent perception, interaction, and reasoning.

\begin{figure*}[t]
\begin{center}
    \includegraphics[width=\linewidth]{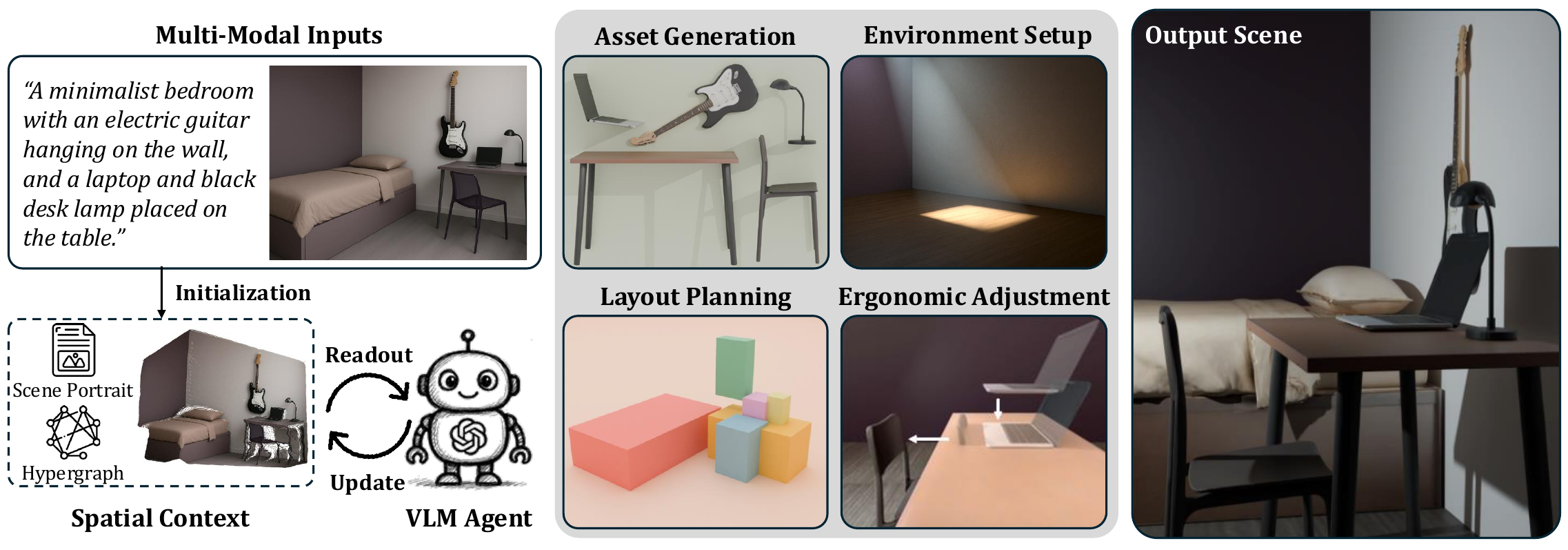}
\end{center}
\caption{\textbf{Left: Spatial context.}  
Given multimodal input from the user, we construct a spatial context that is continuously read and updated by the VLM, effectively injecting it with scene-level semantics, geometry, and relational structure.  
\textbf{Right: Agentic 3D scene generation.}  
Grounded in this context, the VLM performs a four-stage generation process: asset generation, coarse layout planning, environment setup, and ergonomic adjustment—producing a visually coherent and semantically aligned 3D scene.
}
\label{fig:pipeline}
\end{figure*}

To bridge this gap, we propose a framework that \emph{injects spatial context into vision-language models (VLMs)}, integrating their inherent multimodal reasoning capabilities with structured 3D understanding, see \Cref{fig:pipeline}. This context combines multimodal cues to encode an initial understanding of a scene’s semantics, geometry, and layout, providing a grounded representation that informs both 3D scene synthesis and downstream spatial reasoning tasks. Given multimodal input—comprising one or more images, textual descriptions, or both—the spatial context is constructed from three components: a \emph{scene portrait}, which serves as a high-level semantic blueprint through a combination of descriptive text and visual reference; a \emph{semantically labeled point cloud}, produced by a geometric foundation model to capture fine-grained object geometry and spatial layout; and a \emph{scene hypergraph}, which models inter-object relationships. Unlike traditional pairwise scene graphs, our hypergraph formulation captures a broader spectrum of spatial constraints—including unary, binary, and higher-order relations—enabling expressive and ergonomic spatial reasoning~\cite{hypergraph}. Together, these components provide the VLM with a dynamic, multimodal, and geometry-aware context for generating, understanding, and editing coherent 3D scenes.

Building on the spatial context and orchestrated through iterative VLM readout and update, we develop \emph{an agentic generation pipeline} that produces coherent and semantically grounded 3D scenes. To address challenges such as occlusion and limited viewpoints in individual 3D asset generation, we introduce a lightweight \emph{geometric restoration module} that reconstructs complete object geometry from partial observations. To evoke the intended atmosphere and ensure structural and stylistic alignment with the scene’s layout and semantics, in the \emph{environment setup} stage the VLM generates Blender code that constructs the surrounding environment, instantiating architectural elements, terrain, water bodies, and atmospheric effects, augmented by auto-verification against the spatial context. Moreover, leveraging the relational constraints encoded in the scene hypergraph, the VLM performs \emph{ergonomic adjustment} to refine object poses, enforcing physically plausible and semantically meaningful spatial relationships.

In our experiments, comparisons with state-of-the-art methods demonstrate that our framework can generate semantically aligned 3D scenes across a diverse range of challenging inputs—including Chinese poetry, oil paintings, realistic photographs, and even unstructured, unposed image sets. 
Ablation studies further validate the design choices in our pipeline. We also find that, when injected with spatial context, the VLM gains the capacity to support a wide range of downstream tasks, such as interactive scene editing and path planning, implying potential for advancing spatially grounded applications in embodied AI.

In summary, our key contributions are as follows:
\begin{itemize}
\item We propose constructing a continually updatable spatial context and injecting it into VLMs, activating their inherent multimodal reasoning capabilities for structured 3D scene understanding and generation.
    
\item Building on this mechanism, we design an agentic framework that enables 3D scene generation—featuring asset generation with geometric restoration, environment setup through auto-verification against the spatial context, and ergonomic adjustment guided by the scene hypergraph.

\item Our agentic scene generation framework is capable of handling a wide range of challenging inputs—including classical Chinese poetry, oil paintings, and unstructured, unposed image sets—demonstrating a level of generalization that, to our knowledge, no prior method has achieved.

\item Our experiments further show that, with spatial context injection, VLMs gain the ability to perform a range of downstream spatial tasks, including interactive scene editing and path planning.

\end{itemize}

\section{Related Work}

\noindent\textbf{3D scene generation. }
Compared to single-object generation, synthesizing a coherent 3D scene with multiple objects demands both detailed modeling and layout reasoning that balances aesthetic and functional constraints.
Early works \cite{devries2021unconstrained, bautista2022gaudi, chen2023scenedreamer, zhang2023berfscene} employed generative models to learn holistic 3D scene distributions. For example, \cite{liu2021infinite, li2022infinitenature} generated unbounded natural scenes via GAN-based view synthesis, while \cite{hao2021gancraft} translated semantic maps into radiance fields.
More recent approaches leverage 2D diffusion models to synthesize scenes from images or text. Methods such as \cite{NEURIPS2023_7d62a85e, yu2024wonderjourney, hoellein2023text2room, zhang2024text2nerf, li2025dreamscene} iteratively predict 2D content and lift it to 3D via depth estimation. \cite{zhou2025dreamscene360} further extends this to panorama-to-3D conversion. However, these methods typically produce monolithic scene representations, limiting object-level control and editability.
To address this, compositional scene generation has gained traction~\cite{zhai2023commonscenes, epsteindisentangled}. For instance, \cite{Paschalidou2021NEURIPS, po2024compositional} guide generation with layout priors, and \cite{gao2024graphdreamer} and \cite{yang2024holodeck} leverage language models to construct scene graphs or spatial relations. 
ACDC~\cite{dai2024acdc} reduces the cost of generating analogous virtual environments and enhances sim-to-real robustness by constructing a diverse distribution of geometry- and semantics-preserving ``digital cousin'' scenes.
Concurrent to this work, CAST~\cite{yao2025cast} performs component-aligned 3D scene reconstruction from single RGB images, using a GPT-based model for spatial analysis, occlusion-aware 3D generation for object geometry, and physics-aware correction to enforce constraints.
Yet, these approaches often rely on pre-defined 3D assets or fall short in handling fine-grained geometry and complex inter-object relationships.

\noindent\textbf{Layout generation.}
Accurate object placement is essential for compositional 3D scene synthesis, requiring the estimation of positions and orientations that satisfy both functional and aesthetic constraints. Traditional methods \cite{kjolaas2000automatic, coyne2001wordseye, germer2009procedural} relied on rule-based templates or user-defined exemplars \cite{yu2011make}, but often lacked scalability and generalization. Recent data-driven approaches improve robustness by using sequential models \cite{wang2021sceneformer, Paschalidou2021NEURIPS, sun2025forest2seq} or denoising diffusion \cite{para2023cofs, tang2024diffuscene}.
Efforts have also been made to involve LLMs for layout generation from natural language \cite{fu2025anyhome, feng2024layoutgpt}, yet these approaches still rely heavily on exemplars and struggle to interpret user intent dynamically.
Moreover, existing methods rarely account for ergonomic principles, are limited to closed vocabularies, and fall short in capturing higher-order spatial relationships (e.g., symmetry, equidistance) beyond simple pairwise constraints. In contrast, our framework supports open-vocabulary object sets, complex relational reasoning, and ergonomics-aware layout refinement grounded in a scene hypergraph.

\noindent\textbf{LLMs for visual programming}
Large Language Models (LLMs) have demonstrated remarkable zero-shot and few-shot capabilities across a wide range of domains, including mathematics and commonsense reasoning~\cite{brown2020language, ouyang2022training, achiam2023gpt, touvron2023llama, dubey2024llama, team2023gemini}. Recent models further extend this competence by integrating visual inputs, enabling multimodal reasoning across text and images~\cite{alayrac2022flamingo, li2023blip, liu2023llava}. In addition, tool-augmented agents leverage external APIs and visual foundation models to tackle increasingly complex tasks~\cite{schick2023toolformer, wang2024internvid, shen2024hugginggpt, wangvoyager}, including visual code synthesis~\cite{wu2023visual, gupta2023visual, suris2023vipergpt} and multimodal generation or editing~\cite{sharma2024vision, lian2024llmgrounded, wu2024self, feng2024layoutgpt, wang2024genartist, yang2024mastering}.
SceneCraft~\cite{hu2024scenecraft} employs an LLM agent to translate textual prompts into 3D scenes via Blender scripting. While effective for basic compositions, such approaches lack explicit spatial grounding and struggle with high scene complexity, ergonomic constraints, and open-vocabulary object configurations.
In contrast, our work injects a structured \emph{spatial context} into vision-language models, enabling them to maintain a dynamic, geometry-aware internal representation of 3D scenes and handle more complex, semantically rich generation tasks.

% Neurosymbolic Models for Computer Graphics, wu jiajun

\section{Spatially Contextualized VLMs}
Our framework equips the vision-language model (VLM) with a structured spatial context that serves as the backbone of the entire 3D scene generation pipeline. This context integrates multimodal cues to encode an initial understanding of the scene’s semantics, geometry, and layout, providing a grounded representation that informs both scene synthesis and downstream spatial reasoning tasks.

\subsection{Spatial Context Initialization}
Given the user's multimodal input, comprising one or more images, textual descriptions, or their combination, we initialize the spatial context through the following components:

\noindent\textbf{Scene portrait.}  
The VLM first constructs a multimodal scene portrait \( S \), a structured, high-level representation of the scene. This portrait consists of a \emph{detailed textual description} summarizing the scene’s layout, objects, style, atmosphere, and other contextual cues, along with an image—either user-provided or generated from the portrait text as a visual proxy when no image is supplied. Together, these components form a rich blueprint that guides subsequent 3D scene construction and reasoning. 

\noindent\textbf{Semantically labeled point cloud.}  
We employ a geometric foundation model, Fast3R~\cite{Yang_2025_Fast3R}, to generate a colored point cloud from the scene portrait image(s). The resulting point cloud is denoted as \( P = \{ (\mathbf{x}_i, \mathbf{c}_i, l_i) \}_{i=1}^{N} \), where each point \( \mathbf{x}_i \in \mathbb{R}^3 \) has an RGB color \( \mathbf{c}_i \in \mathbb{R}^3 \), and an instance label \( l_i \in \mathbb{N} \), obtained via Grounded-SAM~\cite{ren2024grounded}, which detects object masks on the portrait image(s). For multi-view inputs, object detections are reprojected into 3D and merged based on spatial overlap and semantic similarity. This semantically labeled point cloud provides a spatially grounded and object-centric scaffold for guiding scene construction.

\noindent\textbf{Scene hypergraph.}  
To support layout generation and ergonomic reasoning in 3D scene synthesis, it is essential to model the relationships among object instances within a scene. Recent studies have shown that large language models (LLMs) can effectively interpret and reason over hypergraph structures~\cite{hypergraph}. Inspired by this capability, our approach adopts a hypergraph formulation to represent spatial relationships in complex 3D environments.
From the list of object instances and their corresponding axis-aligned bounding boxes (AABBs) derived from the point cloud \( P \), the VLM constructs a scene hypergraph \( G = (V, E) \), where nodes \( V \) represent object instances, and each hyperedge \( e \in E \) connects one or more nodes to encode spatial relationships. Unlike traditional scene graphs~\cite{Armeni_2019_ICCV}, which are restricted to pairwise relations, our hypergraph formulation naturally captures a broader range of interactions. These include \emph{unary relations}, such as clearance; \emph{binary relations}, such as contact and alignment; and \emph{higher-order relations}, such as equidistance and symmetry.
This component of the spatial context provides the VLM with a flexible and expressive representation of spatial dependencies. 

The complete spatial context \( C = (S, P, G) \) unifies semantic intent, geometric structure, and object-level relationships into a dynamic, temporally evolving representation.

\subsection{Spatial Context Readout and Update}
Unlike static descriptions, the spatial context is iteratively interpreted and updated throughout the scene generation pipeline, allowing the VLM to maintain a grounded and adaptive understanding of the environment.

\noindent\textbf{Readout.}  
To support tasks such as individual asset generation or ergonomics-aware layout refinement, the VLM continuously reads from the spatial context as its primary source of guidance.  
The scene portrait—comprising structured text and images—can be directly interpreted by the VLM through its native multimodal capabilities.  
The scene hypergraph, expressed in a textual format, can likewise be parsed and reasoned over without the need for specialized processing.
The semantically labeled point cloud, however, poses greater challenges for interpretation. Unlike text or images, point clouds and meshes are inherently sparse, unordered, and non-grid-aligned, making them difficult for VLMs to process directly. To address this, we propose projecting the 3D point cloud into 2D RGB+instance point maps. Specifically, we render the point cloud from all available input camera viewpoints, using poses provided by the geometric model~\cite{Yang_2025_Fast3R}. If only a single input view is available, we additionally project the point cloud from canonical orthographic directions—e.g., along the top-down (\(-y\)) and side-view (\(+x\) or \(-x\)) axes—aligned with the scene’s principal orientation, assuming the camera faces the negative \(z\)-axis in a right-handed coordinate system.  
We find that this projected representation preserves sufficient spatial and semantic cues for the VLM to interpret effectively, without requiring native support for raw 3D data.

\noindent\textbf{Update.}  
As the scene evolves, the spatial context is updated on a per-instance basis. When the VLM determines that an object \( v \in V \) requires modification—such as asset replacement or geometric transformation—it retrieves the associated point cloud segment \( P_v \) from the full scene point cloud \( P = \{ (\mathbf{x}_i, \mathbf{c}_i, l_i) \}_{i=1}^N \), where \( \mathbf{x}_i \in \mathbb{R}^3 \) is the 3D coordinate, \( \mathbf{c}_i \in \mathbb{R}^3 \) is the RGB color, and \( l_i \in \mathbb{N} \) is the instance label. The segment \( P_v \subseteq P \) is extracted via masking as \( P_v = \{ (\mathbf{x}_i, \mathbf{c}_i) \mid l_i = v \} \).
Upon obtaining a revised version \( \hat{P}_v \), the global point cloud is updated via \( P \leftarrow (P \setminus P_v) \cup \hat{P}_v \).
This mechanism allows the VLM to incorporate localized changes into the global spatial context, ensuring that all subsequent reasoning and generation steps operate on a coherent and up-to-date world model.

\section{Agentic 3D Scene Generation}
With VLMs injected with spatial context, we propose an agentic framework for 3D scene generation. Specifically, once the spatial context is initialized, the VLM actively engages with it—continuously reading from it to guide generation, and dynamically updating it to reflect scene evolution.
%This section details the full generation pipeline.

\subsection{High-Quality Individual Asset Generation}
The pipeline begins by leveraging the spatial context to identify object instances and synthesize high-quality, individual textured 3D meshes.  
For every object instance \( v \in V \) in the scene hypergraph, the VLM agent retrieves its corresponding point cloud segment \( P_v \subseteq P \), where \( P \) is the global scene point cloud.  
Due to occlusions, limited viewpoints,  artifacts introduced by the geometric foundation model, the retrieved \( P_v \) is often sparse, fragmented, or incomplete—posing a significant challenge for reliable 3D asset synthesis.

\noindent\textbf{Geometric restoration.}  
To overcome these limitations, a lightweight geometric restoration module is employed to reconstruct complete object geometry from partial point cloud observations. Our method builds upon Point-M2AE~\cite{zhang2022point}, with targeted adaptations to accommodate the sparsity patterns observed in Fast3R-generated inputs. To simulate realistic degradation scenarios during training, we randomly occlude regions of complete single-object point clouds and supervise restoration using  uncorrupted shapes.

For each object \( v \), the system first evaluates whether the extracted point segment \( P_v \) is sufficiently complete. If deemed incomplete, the restoration module is applied to produce a densified version \( \hat{P}_v \), and the global point cloud is updated via  
\( P \leftarrow (P \setminus P_v) \cup \hat{P}_v \).  
The resulting instance point cloud is then projected into a canonical front-view image, rasterized onto a 2D viewplane using a fixed virtual camera pose, to generate a clean, front-aligned rendering suitable for mesh generation. This image is subsequently passed to a 3D asset generator, which synthesizes a textured mesh from the projected input.

\subsection{Coarse Layout Planning}
After generating textured meshes for all object instances, we estimate a globally consistent scene arrangement by aligning each mesh with its corresponding point cloud segment in the spatial context.

\noindent\textbf{Optimization objective.}  
Let \( M_v = \{ \mathbf{m}_i \in \mathbb{R}^3 \} \) denote the set of mesh vertices for object \( v \), and let \( P_v = \{ \mathbf{p}_j \in \mathbb{R}^3 \} \) represent the associated point cloud segment.  
The system estimates a similarity transformation—comprising scale \( s \in \mathbb{R}_+ \), rotation \( R \in \mathrm{SO}(3) \), and translation \( \mathbf{t} \in \mathbb{R}^3 \)—by solving:
\begin{equation}
(s^*, R^*, \mathbf{t}^*) = \arg\min_{s, R, \mathbf{t}} \sum_{i} \left\| s R \mathbf{m}_i + \mathbf{t} - \mathrm{NN}_{P_v}(s R \mathbf{m}_i + \mathbf{t}) \right\|^2,
\end{equation}
where \( \mathrm{NN}_{P_v}(\cdot) \) denotes the nearest neighbor in \( P_v \) for a given transformed mesh vertex.

\noindent\textbf{Optimization strategy.}  
The alignment process begins with a coarse initialization: the system translates the mesh to match the centroid of \( P_v \), and then aligns principal axes via oriented bounding box (OBB) fitting. This is followed by a refinement stage using an ICP (Iterative Closest Point) variant to minimize point-to-point distances between the transformed mesh and the target point cloud.
To improve computational efficiency and numerical stability, we apply uniform subsampling to both mesh vertices and point cloud points during each iteration.  
After computing the optimal transformation, the VLM updates the spatial context by replacing the original mesh pose with the refined alignment.

\vspace{-0.1in}

\subsection{Environment Setup with Auto-Verification}
Next, the VLM reasons over the spatial context and generates Blender code to construct the surrounding environment, ensuring structural and stylistic alignment with the scene’s layout and semantics.

\noindent\textbf{For indoor scenes,} environment setup instantiates architectural elements such as walls, floors, and ceilings, with specified geometry, placement, materials, and textures. The VLM integrates these elements into the spatial context by adding corresponding vertices and hyperedges to the scene hypergraph and extending the point cloud with samples from the generated geometry.  
It also configures interior lighting by selecting appropriate source types (e.g., point, area, or spot) and adjusting parameters such as intensity and color.

\noindent\textbf{For outdoor scenes,} VLM generates environmental components e.g., sky domes with sky textures to simulate daylight and atmosphere, terrain surfaces constructed via procedural terrain generators to introduce natural topography, bodies of water created using displacement and wave modifiers to mimic surface undulation, and volumetric effects (e.g., fog or haze) implemented through the Principled Volume shader, with carefully tuned density and anisotropy parameters to control light scattering and depth perception.

\noindent\textbf{Auto-verification against spatial context.}  
Despite VLM's strong visual programming capabilities, directly authoring Blender code for environment setup remains challenging—even when guided by our proposed spatial context.
Thus, we introduce an \textit{auto-verification procedure} that enables the VLM to self-check the consistency of its generated code. After producing the initial environment code, the system renders an image of the resulting scene and performs self-evaluation using a chain-of-thought reasoning process to identify inconsistencies between the rendered output and the expected spatial context. Based on this analysis, the VLM then refines the code to correct identified issues.
We find that this iterative verification-and-refinement loop significantly improves semantic and structural alignment with the spatial context, while also reducing the frequency of rendering errors and unintended artifacts.

\subsection{Hypergraph-based Ergonomic Adjustment}
\label{sec:ergo}

\begin{figure*}[t]
\begin{center}
    \includegraphics[width=0.98\linewidth]{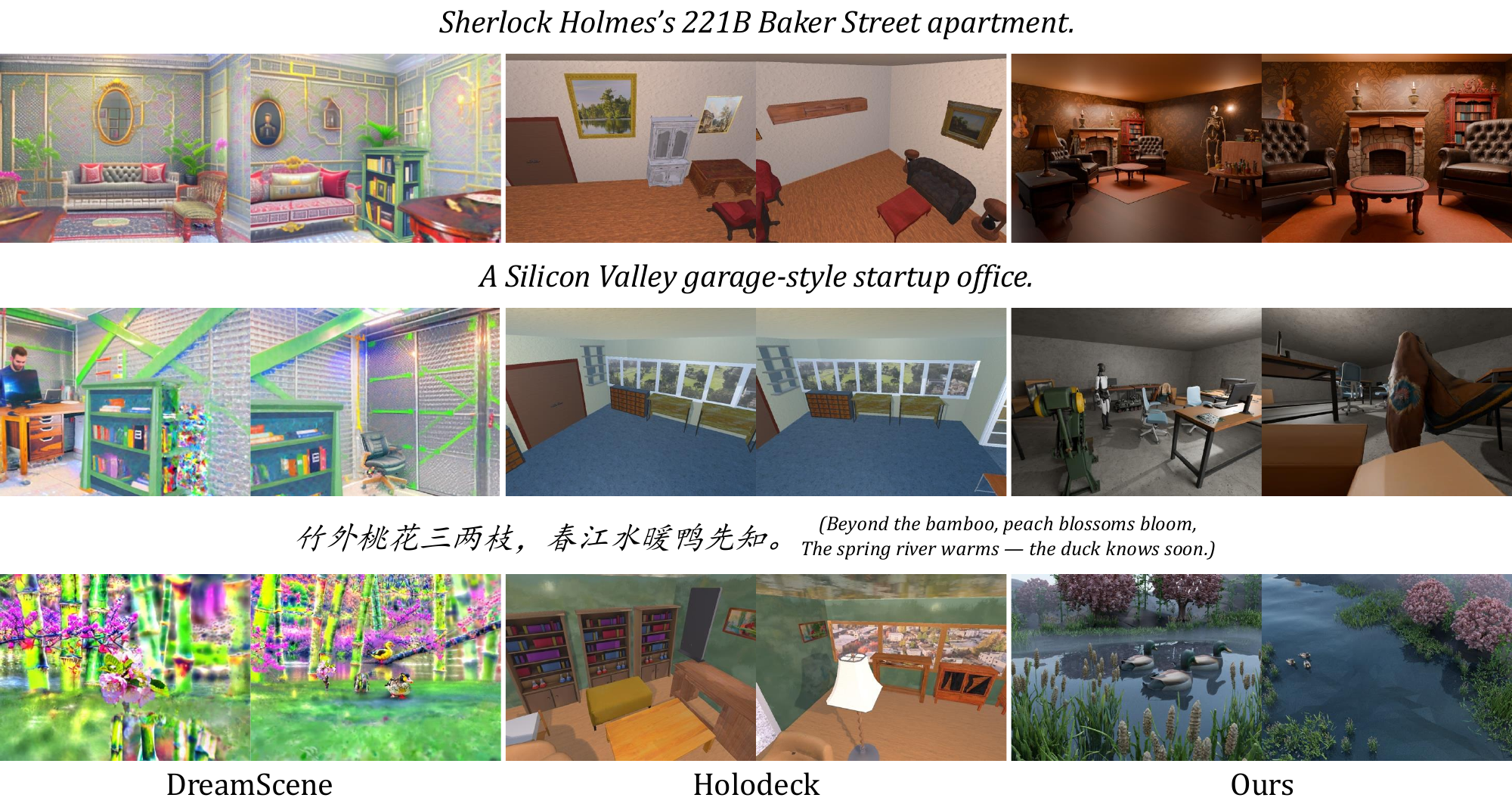}
\end{center}

\caption{\textbf{Qualitative comparison for text-based 3D scene generation.}
Our method produces more coherent, stylistically aligned, and visually plausible scenes compared to DreamScene~\cite{li2025dreamscene} and Holodeck~\cite{yang2024holodeck}.}
\label{fig:comparison_text}

\end{figure*}

While the initial layout generation places each individual asset in a globally consistent position based on the spatial context, it often results in structural issues such as inter-object penetration, detachment, or misalignment with ergonomic expectations. To address these, the VLM performs a joint optimization over object poses to refine the overall arrangement and enforce physically and functionally meaningful spatial relations.
We optimize the object transformations \( \{ R_v, \mathbf{t}_v \}_{v \in V} \) to satisfy soft spatial constraints encoded in the scene hypergraph \( G = (V, E) \). Each hyperedge \( e \in E \) corresponds to a spatial relation type \( r_e \in R \), where \( R = \{\text{clearance}, \text{contact}, \text{alignment}, \text{equidistance}, \text{symmetry} \} \). These cover unary (clearance), binary (contact, alignment), and ternary (equidistance, symmetry) relationships.
The optimization objective is:
\begin{equation}
\min_{\{ R_v, \mathbf{t}_v \}_{v \in V}} \sum_{e \in E} \lambda_{r_e} \cdot L_{r_e}(\{ R_v, \mathbf{t}_v \}_{v \in e}),
\end{equation}
where \( L_{r_e} \) is relation-specific loss and \( \lambda_{r_e} \) is its associated weight.

\noindent\textbf{Relation-specific loss.}  
We use the contact relation as a representative example; definitions of the remaining losses are provided in \Cref{app:ergonomic-loss}.  
To encourage physical contact between two objects \( v_i \) and \( v_j \), we minimize the distance between their closest transformed surface points. Let \( M_{v_i} \) and \( M_{v_j} \) be sampled surface points. After transformation, the points become \( \tilde{\mathbf{p}} = R_{v_i} \mathbf{p} + \mathbf{t}_{v_i} \) and \( \tilde{\mathbf{q}} = R_{v_j} \mathbf{q} + \mathbf{t}_{v_j} \). The contact loss is:
\begin{equation}
L_{\text{contact}} = \left[ \min_{\mathbf{p}, \mathbf{q}} \left\| \tilde{\mathbf{p}} - \tilde{\mathbf{q}} \right\| - \epsilon \right]_+^2,
\end{equation}
where \( [\cdot]_+ = \max(0, \cdot) \), and \( \epsilon \) is a small soft contact margin.

\noindent\textbf{Soft constraints configuration.}  
Some relation-specific losses require VLM to determine constraint details through contextual reasoning. Our spatial context provides the necessary semantic and geometric cues for VLM to infer which axes to align, where to enforce contact, and how much clearance is appropriate.  
Once optimized, the transformations \( \{ R_v, \mathbf{t}_v \}_{v \in V} \) are applied to update the spatial context by repositioning each instance mesh to its final pose.

\begin{figure*}[t]
\begin{center}
    \includegraphics[width=0.98\linewidth]{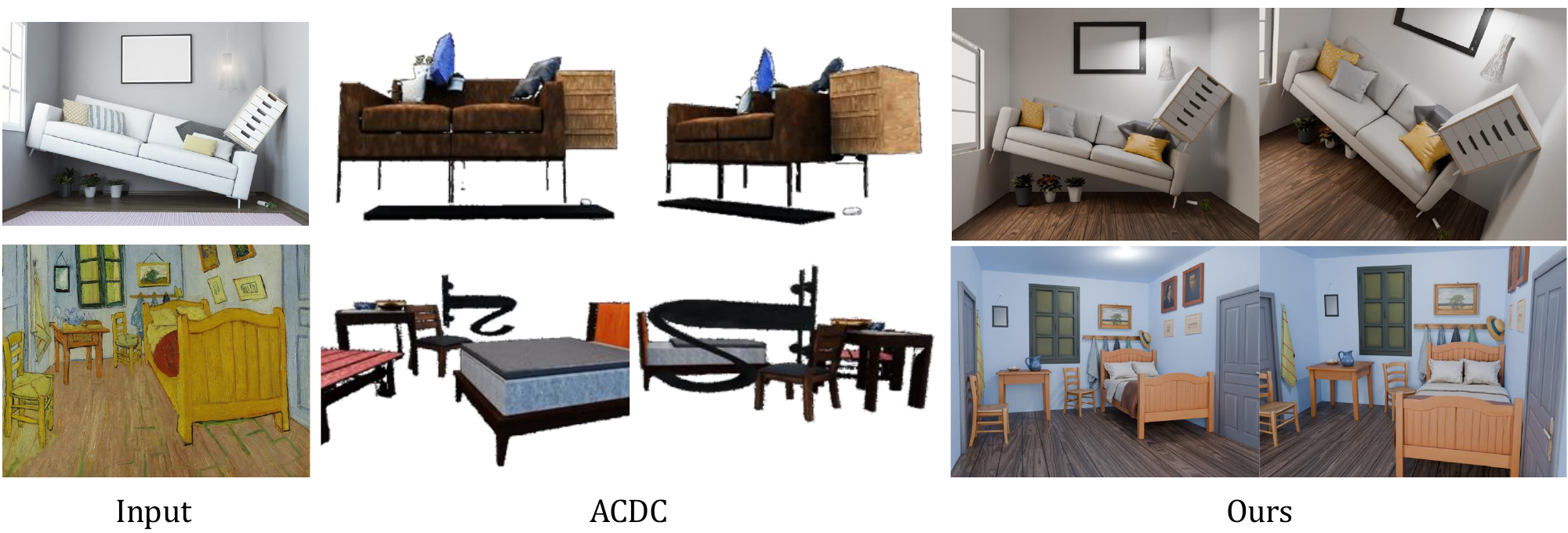}
\end{center}

\caption{\textbf{Qualitative comparison for image-based 3D scene generation.}
Compared to ACDC~\cite{dai2024acdc}, our method appears to generate scenes that more consistently reflect the spatial and visual characteristics of the input images.}
\label{fig:comparison_image}

\end{figure*}

\section{Experiments}
We evaluate our proposed framework for 3D scene generation across a diverse set of challenging scenarios. Our experiments include comparisons with state-of-the-art baselines and ablation studies to validate the effectiveness of key components. We further demonstrate the capabilities of the spatially contextualized VLM in performing downstream spatially grounded tasks. For additional results and implementation details, please refer to our \emph{figures-only pages, supplementary material, and accompanying video}.

\noindent\textbf{Implementation details.}  
We adopt GPT-4o~\cite{achiam2023gpt} as the VLM integrating the spatial context and acting as the agent throughout the 3D scene generation pipeline. \emph{Prompts used to construct the spatial context are provided in the appendix.}
Our geometric restoration module is trained on point maps estimated by Fast3R~\cite{Yang_2025_Fast3R} using the CO3D~\cite{reizenstein21co3d} training images. The model converges in approximately 3 hours on an NVIDIA A100 GPU. During asset generation, we use the Meshy API\footnote{\url{https://www.meshy.ai/api}} for image-to-3D synthesis. For layout planning and ergonomic adjustment, optimization problems are implemented using PyTorch. 
All final 3D scenes are rendered using the Blender Cycles rendering engine to produce photorealistic results with accurate lighting and material representation.

\noindent\textbf{Metrics.} 
\textit{(i) Geometric fidelity.} 
We use Chamfer Distance (CD), which averages two terms: accuracy (the smallest Euclidean distance from reconstructed shape points to ground-truth points) and completeness (the smallest Euclidean distance from ground-truth points to reconstructed shape points).
\textit{(ii) Instance overlap.} 
We compute Intersection over Union (IoU) to measure instance-level overlap between reconstructed and ground-truth scenes.
\textit{(iii) Semantic alignment.} 
To assess alignment with input prompts, we render images from synthesized scenes and compute text-image similarity using \emph{CLIP}~\cite{clip} and \emph{BLIP}~\cite{li2023blip}, and image-image similarity using \emph{LPIPS} (AlexNet)~\cite{zhang2018perceptual}.
\textit{(iv) Aesthetic quality and functional plausibility.} 
We evaluate aesthetic quality (AQ) and functional plausibility (FP) through human ratings from a user study with 16 participants and GPT-4o ratings. Methods are ranked based on averaged ordinal scores across a benchmark set of scenes, with lower ranks indicating better performance.

\begin{table}[t]
\centering
\caption{Quantitative comparison of semantic alignment (CLIP, BLIP, LPIPS), aesthetic quality (AQ), and functional plausibility (FP).}
\label{tab:comparison}
\resizebox{\linewidth}{!}{
\begin{tabular}{lccccc}
\toprule
\textbf{Method} & \textbf{CLIP ($\uparrow$)} & \textbf{BLIP ($\uparrow$)} & \textbf{LPIPS ($\downarrow$)} & \textbf{AQ (4o/User) ($\downarrow$)} & \textbf{FP (4o/User) ($\downarrow$)} \\
\midrule
Holodeck        & 0.274 & 0.461 & - & 3.00 / 3.25 &  3.00 / 2.69 \\
DreamScene      & 0.219 & 0.509 & - & 4.00 / 2.75 &  4.00 / 2.75 \\
ACDC            & - & - & 0.760 & 2.00 / 2.94 &  2.00 / 3.31 \\
\textbf{Ours}   & \textbf{0.385} & \textbf{0.737} & \textbf{0.571} & \textbf{1.00} / \textbf{1.06} & \textbf{1.00} / \textbf{1.19} \\
\bottomrule
\end{tabular}
}

\end{table}

\begin{table}[t]
\centering
\caption{Quantitative comparison of geometric fidelity and instance overlap on 3D-FRONT dataset~\cite{3dfront,3dfuture}, and evaluation of the impact of the number of input view.}
\label{tab:comparison_3dfront}
\resizebox{0.9\linewidth}{!}{
\begin{tabular}{lccccc}
\toprule
\textbf{Method} & \textbf{\#Views} & \textbf{Acc. ($\downarrow$)} & \textbf{Comp. ($\downarrow$)} & \textbf{CD ($\downarrow$)} & \textbf{IoU ($\uparrow$)} \\
\midrule
ACDC       & 1  & 17.32 & 8.51  & 12.915 & 52.4 \\
{Ours} &  1  & \textbf{7.73} & \textbf{2.51} & \textbf{5.12} & \textbf{70.8} \\
\midrule
{Ours}& 5   & {4.16} & {2.62} & {3.39} & {75.4} \\
{Ours} & 10  & {1.89} & {0.97} & {1.43} & {78.1} \\
{Ours} & 20  & \textbf{1.61} & \textbf{0.93} & \textbf{1.27} & \textbf{78.8} \\
\bottomrule
\end{tabular}
}
\end{table}

\begin{figure*}[t]
\begin{center}
\includegraphics[width=\linewidth]{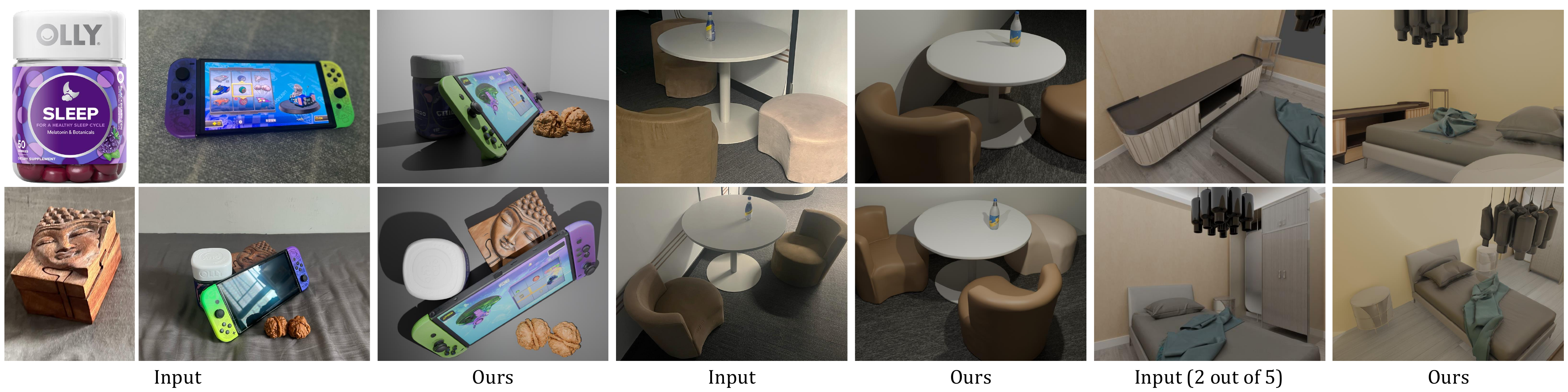}
\end{center}
\caption{\textbf{Results from multi-view observations.} Our method synthesizes consistent scenes from unposed, unstructured image collections.}
\label{fig:multi}
\end{figure*}

\begin{figure}[t]
\begin{center}
    \includegraphics[width=1.0\linewidth]{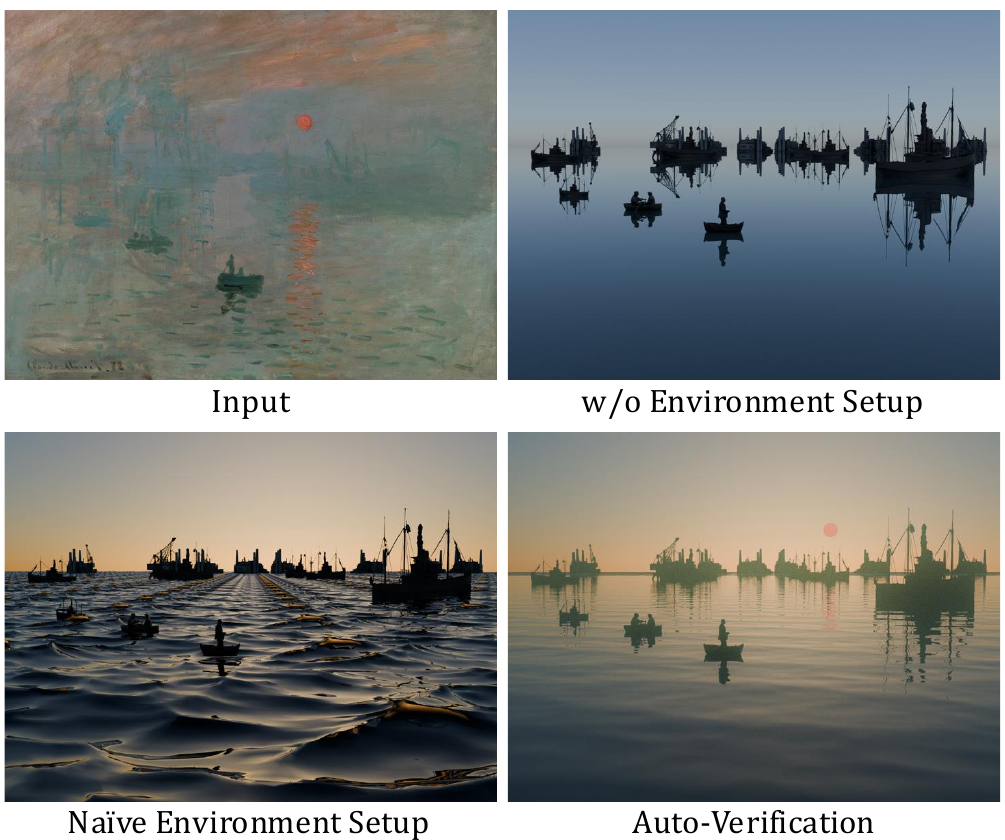}
\end{center}

\caption{\textbf{Ablation on environment setup}. Without structured setup, scenes lack realistic lighting and environmental elements. Naïve modifiers yield low-fidelity results, while our auto-verified setup produces coherent, atmospheric environments aligned with spatial context.
}

\label{fig:env}
\end{figure}

\subsection{Comparison}

\noindent\textbf{Text-conditioned generation.}
We compare our framework against two recent text-to-3D methods—Holodeck~\cite{yang2024holodeck} and DreamScene~\cite{li2025dreamscene}. As shown in \Cref{fig:comparison_text}, our method produces scenes that more faithfully preserve semantic alignment, spatial structure, and stylistic intent. For example, in the \emph{Holmes apartment} case, our result better captures the Victorian layout and furniture arrangement, while others exhibit geometric artifacts or overlook contextual cues. Quantitatively, our method achieves the highest CLIP and BLIP scores in \Cref{tab:comparison}, reflecting superior consistency with input prompts. It also ranks best in aesthetic quality (AQ) and functional plausibility (FP), based on both GPT-4o and user evaluations.

\noindent\textbf{Image-conditioned generation.}
\Cref{fig:comparison_image} shows a comparison with ACDC~\cite{dai2024acdc}, a recent method for real-to-sim scene construction. 
Our system more effectively reconstructs spatial layouts and scene compositions, such as the tilted sofa in a living room, while better preserving the stylistic integrity of iconic works like Van Gogh’s  \emph{Bedroom in Arles}.
In \Cref{tab:comparison}, our approach achieves the best image-image similarity score, demonstrating higher visual fidelity to the input images. For geometric accuracy and alignment at the instance level with ground truth, \Cref{tab:comparison_3dfront} shows that our approach achieves a lower Chamfer distance and higher IoU.

\noindent\textbf{Image set as input.}
Unlike prior methods, which are typically restricted to single-view input, our framework naturally accommodates unstructured and unposed image collections. As illustrated in \Cref{fig:multi}, our system consolidates geometric cues from diverse viewpoints into a coherent 3D layout. This ability stems from the VLM's integration with our spatial context, which provides a flexible representation for resolving spatial correspondences across views. We also evaluate the impact of the number of input views on performance. As shown in \Cref{tab:comparison_3dfront}, increasing the number of views improves the precision of the reconstruction.

\begin{figure}[t]
\begin{center}
    \includegraphics[width=0.75\linewidth]{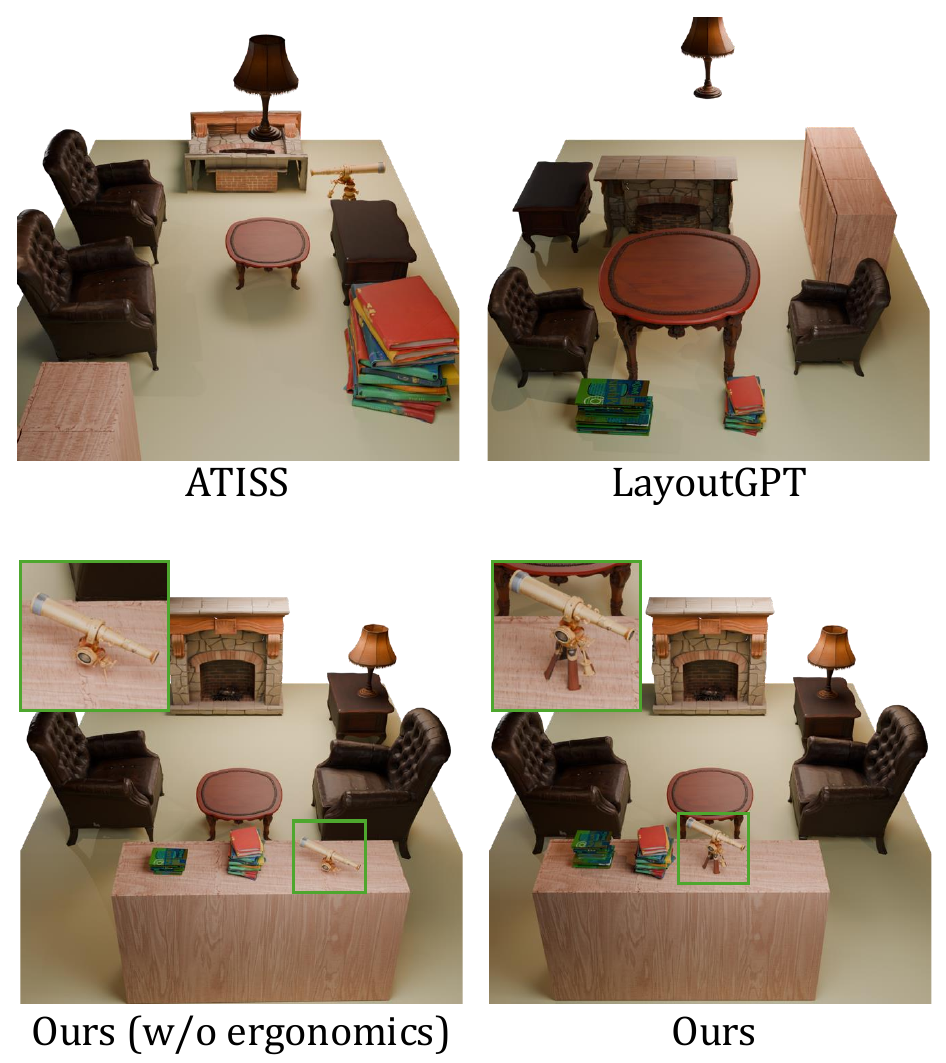}
\end{center}
\caption{\textbf{Ablation on layout planning and ergonomic adjustment.} Compared to ATISS~\cite{Paschalidou2021NEURIPS} and LayoutGPT~\cite{feng2024layoutgpt}, our layout preserves scale and placement accuracy. Removing ergonomic adjustment results in collisions and misalignment.
}
\label{fig:layout}
\end{figure}

\subsection{Ablation Study}

\noindent\textbf{Environment Setup.}
We evaluate the importance of environment setup and the role of auto-verification. As shown in \Cref{fig:env}, without this module, key visual elements—such as sky texture, sunlight, or water surfaces—are either missing or appear unnatural. Introducing a naïve environment setup with basic modifiers (e.g., for water) adds some structure, but the results often lack realism—waves may appear flat or physically implausible. In contrast, our auto-verified environment setup significantly enhances scene realism and atmosphere by ensuring alignment with the spatial context and refining visual fidelity through iterative code correction.

\noindent\textbf{Layout Planning.}
We assess layout planning by replacing our method with ATISS~\cite{Paschalidou2021NEURIPS} and LayoutGPT~\cite{feng2024layoutgpt}. As shown in \Cref{fig:layout}, these alternatives often introduce scale or placement errors (e.g., floating lamps, misaligned furniture), whereas our method yields more structurally accurate and semantically coherent layouts.
Removing ergonomic adjustment results in object misalignment and interpenetration, leading to degraded visual aesthetics and functional plausibility. These findings highlight the necessity of our ergonomic refinement step for ensuring realistic and usable 3D scenes.

\subsection{Spatially Grounded Downstream Tasks}
\begin{figure}[t]
\begin{center}
\includegraphics[width=1.0\linewidth]{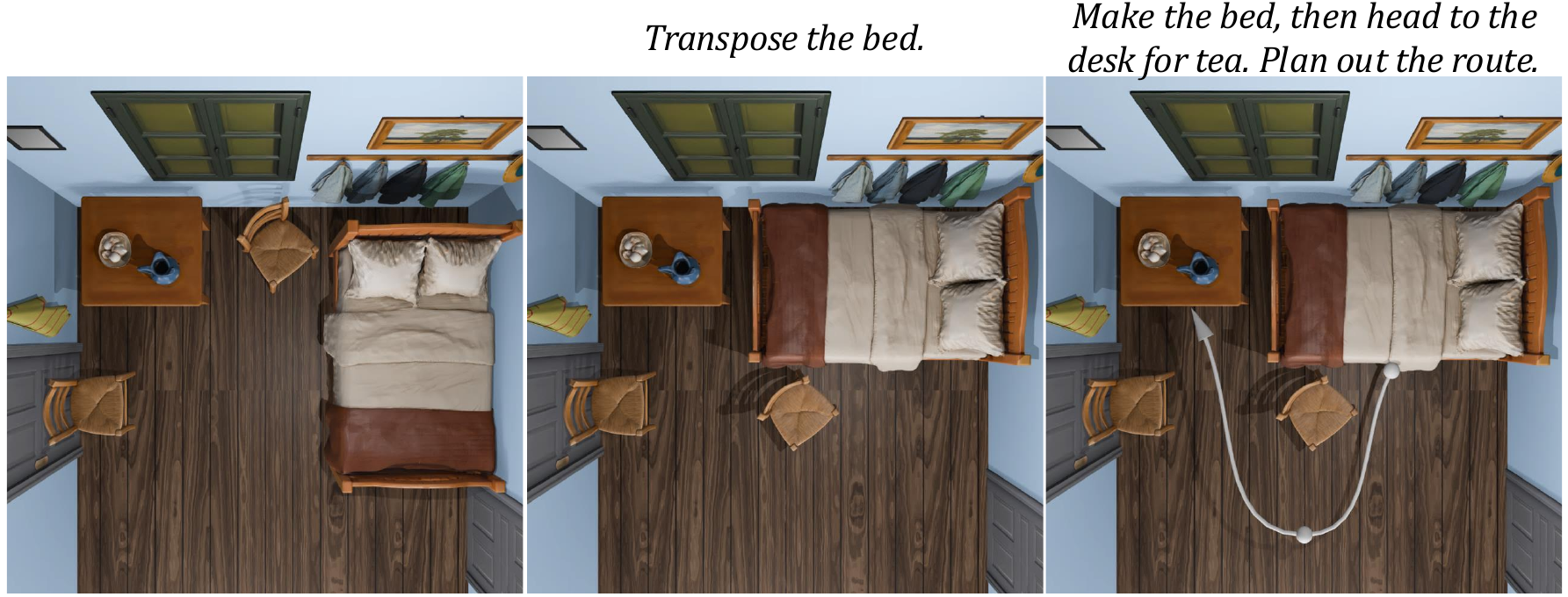}
\end{center}

\caption{\textbf{Scene editing and spatial reasoning.} Our method enables downstream spatial tasks such as furniture manipulation and obstacle-aware path planning, by reasoning over the spatial context.}

\label{fig:editing}
\end{figure}

Our framework supports downstream spatial tasks such as object manipulation and navigation planning. As shown in \Cref{fig:editing}, the VLM can follow high-level instructions—like relocating furniture or planning a route. Notably, it can generate a collision-free path from the bed to the desk without explicit labels or obstacle maps, by implicitly understanding spatial layout and avoiding objects such as the bedside chair. This is enabled by our structured spatial context, which encodes object geometry and relations, and is dynamically updated after editing, allowing the VLM to extract feasible trajectories from the modified scene.

\section{Conclusion}
We present a novel framework that equips vision-language models with structured spatial context. By integrating a scene portrait, a semantically labeled point cloud, and a scene hypergraph, our method provides the VLM with a dynamic, geometry-aware representation for spatial reasoning. Built upon this foundation, our agentic generation pipeline, featuring high-quality asset creation, context-aware environment setup with auto-verification, and ergonomic layout refinement.
Extensive experiments demonstrate that our system generalizes well to diverse and challenging inputs and outperforms existing baselines in scene fidelity and functional coherence. Moreover, the spatial context injection enables VLMs to execute downstream spatial tasks, such as editing and navigation, illustrating their promise for real-world embodied and interactive 3D applications.

{
\small
\bibliographystyle{ieeenat_fullname}

\def\CVPR{IEEE/CVF Conference on Computer Vision and Pattern Recognition (CVPR)}\def\ECCV{ European Conference on Computer Vision (ECCV)}\def\ICCV{IEEE/CVF International Conference on Computer Vision (ICCV)}\def\NIPS{Advances in Neural Information Processing Systems (NeurIPS)}\def\ICML{International Conference on Machine Learning (ICML)}\def\ICLR{International Conference on Learning Representations (ICLR)}\def\WACV{IEEE/CVF Winter Conference on Applications of Computer Vision (WACV)}\def\CVPRW{IEEE/CVF Conference on Computer Vision and Pattern Recognition (CVPR) Workshops}\def\ICCVW{IEEE/CVF International Conference on Computer Vision (ICCV) Workshops}\def\ICRA{IEEE International Conference on Robotics and Automation (ICRA)}\def\TOG{ACM Transactions on Graphics (TOG)}\def\PAMI{IEEE Transactions on Pattern Analysis and Machine Intelligence (PAMI)}\def\TIP{IEEE Transactions on Image Processing (TIP)}\def\IJCV{International Journal of Computer Vision (IJCV)}\def\SIGGRAPH{ACM Transactions on Graphics
  (SIGGRAPH)}\def\SIGGRAPHASIA{ACM Transactions on Graphics (SIGGRAPH Asia)}\def\TOG{ACM Transactions on Graphics (TOG)}\def\threedv{International Conference on 3D Vision (3DV)}\def\TVCG{IEEE Transactions on Visualization and Computer Graphics (TVCG)}\def\PMLR{Proceedings of Machine Learning Research (PMLR)}

}
% \bibliographystyle{ACM-Reference-Format}
% \bibliography{sample-bibliography}

\newpage
\appendix
\onecolumn

\section{Limitations and Future Work}
While our framework demonstrates strong generalization and performance, several limitations remain. First, when the number of object instances is large or includes extremely small objects, spatial context construction may miss instances or introduce noise, potentially affecting layout quality and scene completeness. Second, in the multi-image setting, performance heavily relies on the geometric foundation model used to estimate depth and structure—failure cases in depth prediction can lead to misalignment in the resulting scene. Finally, our current scene hypergraph models unary, binary, and ternary spatial relations; extending this structure to support richer or learned higher-order relations could further enhance ergonomic reasoning and compositional flexibility. Addressing these challenges offers promising directions for future work.

\section{Ergonomic Adjustment: Relation-Specific Constraints}
\label{app:ergonomic-loss}
In this section, we detail the definitions of other relation-specific loss functions used in our ergonomic adjustment module, as referenced in Section 4.4. While the main text introduces the contact constraint, our scene hypergraph formulation supports a richer set of spatial relations—including unary (e.g., clearance), binary (e.g., alignment), and ternary (e.g., symmetry, equidistance). Each is encoded as a soft differentiable loss to guide physically plausible and semantically meaningful spatial arrangements. Below, we present the mathematical formulation and intuition behind each additional constraint type.

\noindent\textbf{Clearance.}  
To prevent spatial crowding and ensure functional space around objects, we introduce a unary clearance constraint that enforces a minimum separation between each object and all others in the scene.  
Let \( \mathbf{o}_{v} \) denote the center of the axis-aligned bounding box (AABB) of object \( v \) in its local frame. After transformation, its world-space position is \( \tilde{\mathbf{o}}_{v} = R_{v} \mathbf{o}_{v} + \mathbf{t}_{v} \).  
For each object \( v \in V \), the clearance loss is defined as:
\begin{equation}
L_{\text{clearance}}(R_v, \mathbf{t}_v) = \sum_{\substack{v' \in V \\ v' \neq v}} \left[ d_{\text{min}}(v) - \left\| \tilde{\mathbf{o}}_{v} - \tilde{\mathbf{o}}_{v'} \right\| \right]_+^2,
\end{equation}
where \( d_{\text{min}}(v) \) is a VLM-determined minimum clearance radius for object \( v \), typically computed from its bounding box size or semantic role, and \( [\cdot]_+ = \max(0, \cdot) \) denotes the hinge function.

\noindent\textbf{Alignment.}  
To promote symmetric or functional alignment between two objects \( v_i \) and \( v_j \)—such as centering a chair relative to a desk—we impose a soft constraint that minimizes their displacement along contextually relevant axes. Let \( \mathbf{o}_{v_i} \) and \( \mathbf{o}_{v_j} \) denote the centers of the axis-aligned bounding boxes (AABBs) of the respective meshes. After applying transformations, the world-space centers become \( \tilde{\mathbf{o}}_{v_i} = R_{v_i} \mathbf{o}_{v_i} + \mathbf{t}_{v_i} \) and \( \tilde{\mathbf{o}}_{v_j} = R_{v_j} \mathbf{o}_{v_j} + \mathbf{t}_{v_j} \). The alignment loss is defined as:
\begin{equation}
L_{\text{align}}(R_{v_i}, \mathbf{t}_{v_i}, R_{v_j}, \mathbf{t}_{v_j}) = 
\left\| \mathbf{A}_{r_{ij}} \left( \tilde{\mathbf{o}}_{v_i} - \tilde{\mathbf{o}}_{v_j} \right) \right\|^2,
\end{equation}
where \( \mathbf{A}_{r_{ij}} \in \mathbb{R}^{d \times 3} \) is a projection matrix that selects the axis or axes relevant to the alignment relation \( r_{ij} \). This encourages alignment along those axes while allowing flexibility in other directions.

\noindent\textbf{Symmetry.}  
To encourage symmetric spatial arrangements, we introduce a ternary symmetry constraint. It ensures that two objects \( v_i \) and \( v_j \) are symmetrically positioned with respect to a reference object \( v_k \) along a contextually relevant axis. The axis of symmetry—typically one of the global \( x \), \( y \), or \( z \) axes—is determined by the VLM based on semantic roles or scene structure.
Let \( \tilde{\mathbf{o}}_{v} = R_{v} \mathbf{o}_{v} + \mathbf{t}_{v} \) denote the transformed AABB center of object \( v \in \{v_i, v_j, v_k\} \).
Let \( \mathbf{A}_{r} \in \mathbb{R}^{1 \times 3} \) be the axis selector vector corresponding to the symmetry relation \( r \in \{x, y, z\} \), e.g., \( \mathbf{A}_x = [1, 0, 0] \). The symmetry loss is defined as:
\begin{equation}
L_{\text{symmetry}} = \left\| \mathbf{A}_{r} \left( \frac{\tilde{\mathbf{o}}_{v_i} + \tilde{\mathbf{o}}_{v_j}}{2} - \tilde{\mathbf{o}}_{v_k} \right) \right\|^2,
\end{equation}
which penalizes deviation of the midpoint between \( v_i \) and \( v_j \) from the center of \( v_k \) along the symmetry axis.

\noindent\textbf{Equidistance.}
To enforce symmetric spacing, we introduce an equidistance constraint where two objects \( v_i \) and \( v_j \) are encouraged to maintain equal distance from a reference object \( v_k \) along a specified axis. Let \( \tilde{\mathbf{o}}_v = R_v \mathbf{o}_v + \mathbf{t}_v \) denote the transformed AABB center for each \( v \in \{v_i, v_j, v_k\} \), and let \( \mathbf{a} \in \mathbb{R}^3 \) be a unit vector representing the axis of comparison. The equidistance loss is defined as:
\begin{equation}
L_{\text{equi}} = \left\| \mathbf{a}^\top (\tilde{\mathbf{o}}_{v_i} - \tilde{\mathbf{o}}_{v_k}) - \mathbf{a}^\top (\tilde{\mathbf{o}}_{v_j} - \tilde{\mathbf{o}}_{v_k}) \right\|^2.
\end{equation}
This loss encourages \( v_i \) and \( v_j \) to be placed symmetrically with respect to \( v_k \) along axis \( \mathbf{a} \).

\section{Addition Qualitative Results}
\begin{figure*}[t]
\begin{center}
\includegraphics[width=0.95\linewidth]{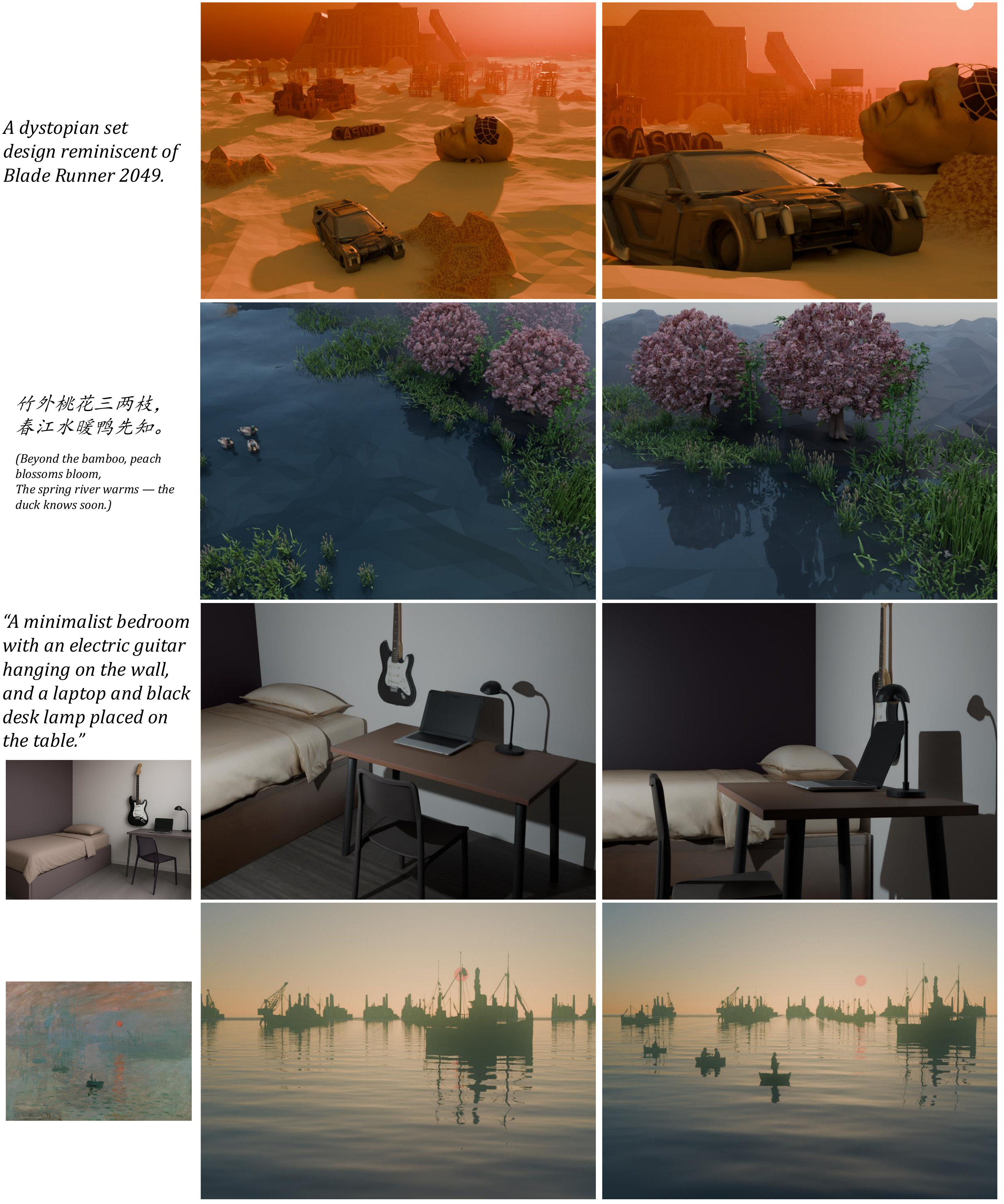}
\end{center}
\caption{\textbf{Additional qualitative results.} }
\label{fig:additional}
\end{figure*}

\end{document}